\documentclass[runningheads]{llncs}
\usepackage{graphicx}

\usepackage{times}
\usepackage{latexsym}
\usepackage{tabularx}
\usepackage{multirow}

\usepackage[hyphens]{url}
\usepackage[utf8]{inputenc}

\begin{document}
\title{A Parallel Corpus of Theses and Dissertations Abstracts}
\author{Felipe Soares\inst{1,2} \and
Gabrielli Harumi Yamashita\inst{2} \and
Michel Jose Anzanello\inst{2}}
\authorrunning{F. Soares et al.}

\institute{Instituto de Informática - UFRGS - Porto Alegre - Brazil \and
Escola de Engenharia - UFRGS - Porto Alegre - Brazil \\
\email{felipe.soares@inf.ufrgs.br, gabrielli.hy@gmail.com, anzanello@producao.ufrgs.br}\\ }

\maketitle              
\begin{abstract}
In Brazil, the governmental body responsible for overseeing and coordinating post-graduate programs, CAPES, keeps records of all theses and dissertations presented in the country. Information regarding such documents can be accessed online in the Theses and Dissertations Catalog (TDC), which contains abstracts in Portuguese and English, and additional metadata. Thus, this database can be a potential source of parallel corpora for the Portuguese and English languages. In this article, we present the development of a parallel corpus from TDC, which is made available by CAPES under the open data initiative. Approximately 240,000 documents were collected and aligned using the Hunalign tool. We demonstrate the capability of our developed corpus by training Statistical Machine Translation (SMT) and Neural Machine Translation (NMT) models for both language directions, followed by a comparison with Google Translate (GT). Both translation models presented better BLEU scores than GT, with NMT system being the most accurate one. Sentence alignment was also manually evaluated, presenting an average of 82.30\% correctly aligned sentences. Our parallel corpus is freely available in TMX format, with complementary information regarding document metadata.

\end{abstract}
\section{Introduction}
\label{intro}

The availability of cross-language parallel corpora is one of the basis of current Statistical and Neural Machine Translation systems (e.g. SMT and NMT). Acquiring a high-quality parallel corpus that is large enough to train MT systems, specially NMT ones, is not a trivial task, since it usually demands human curating and correct alignment. In light of that, the automated creation of parallel corpora from freely available resources is extremely important in Natural Language Processing (NLP), enabling the development of accurate MT solutions. Many parallel corpora are already available, some with bilingual alignment, while others are multilingually aligned, with 3 or more languages, such as Europarl \cite{koehn2005europarl}, from the European Parliament, JRC-Acquis \cite{steinberger2006jrc}, from the European Commission, OpenSubtitles \cite{zhang2014dual}, from movies subtitles. \par

The extraction of parallel sentences from scientific writing can be a valuable language resource for MT and other NLP tasks. The development of parallel corpora from scientific texts has been researched by several authors, aiming at translation of biomedical articles \cite{wu2011statistical,NEVES16.800}, or named entity recognition of biomedical concepts \cite{kors2015multilingual}. Regarding Portuguese/English and English/Spanish language pairs, the FAPESP corpus \cite{aziz:2011:newfapesp}, from the Brazilian magazine \textit{revista pesquisa FAPESP}, contains more than 150,000 aligned sentences per language pair, constituting an important language resource.\par

In Brazil, the governmental body responsible for overseeing post-graduate programs across the country, called CAPES, tracks every enrolled student and scientific production. In addition, CAPES maintains a freely accessible database of theses and dissertations produced by the graduate students (i.e. Theses and Dissertations Catalog - TDC) since 1987, with abstracts available since 2013. Under recent governmental efforts in data sharing, CAPES made TDC available in CSV format, making it easily accessible for data mining tasks. Recent data files, from 2013 to 2016, contain valuable information for NLP purposes, such as abstracts in Portuguese and English, scientific categories, and keywords. Thus, TDC can be an important source of parallel Portuguese/English scientific abstracts. \par

In this work, we developed a sentence aligned parallel corpus gathered from CAPES TDC comprised of abstracts in English and Portuguese spanning the years from 2013 to 2016. In addition, we included metadata regarding the respective theses and dissertations.

%
%
    %
    %
    %
    %
    %
    %

\section{Material and Methods}
\begin{figure}[!h]
\begin{center}
\includegraphics[scale=0.5]{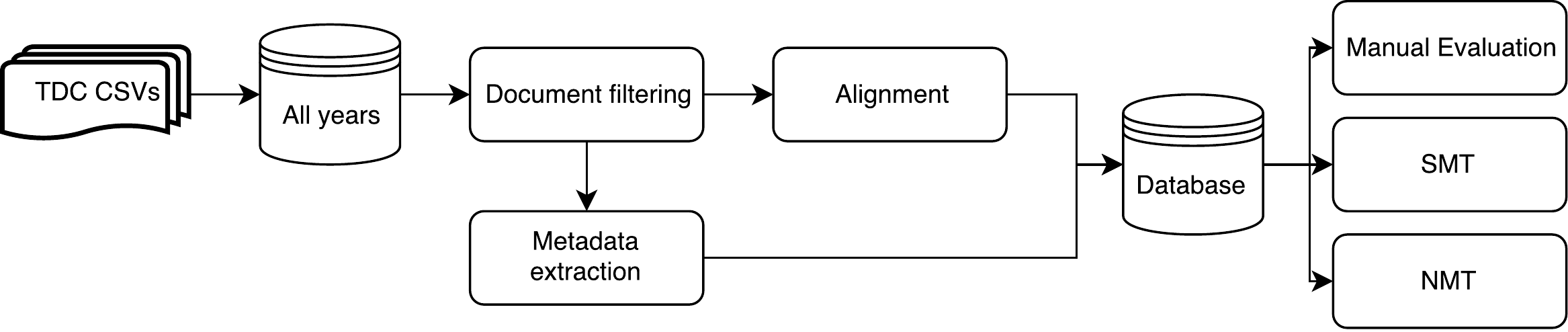} 
\caption{Steps employed in the development of the parallel corpora.}
\label{fig.1}
\end{center}
\end{figure}

In this section, we detail the information retrieved from CAPES website, the filtering process, the sentence alignment, and the evaluation experiments. An overview of the steps employed in this article is shown in Figure \ref{fig.1}.

\subsection{Document retrieval and parsing}

The TDC datasets are available in the CAPES open data website\footnote{\url{https://dadosabertos.capes.gov.br/dataset/catalogo-de-teses-e-dissertacoes-de-2013-a-2016}} divided by years, from 2013 to 2016 in CSV and XLSX formats. We downloaded all CSV files from the respective website and loaded them into an SQL database for better manipulation. The database was then filtered to remove documents without both Portuguese and English abstracts, and additional metadata selected.\par

After the initial filtering, the resulting documents were processed for language checking\footnote{\url{https://github.com/Mimino666/langdetect}} to make sure that there was no misplacing of English abstracts in the Portuguese field, or the other way around, removing the documents that presented such inconsistency. We also performed a case folding to lower case letters, since the TDC datasets present all fields with uppercase letters. In addition, we also removed newline/carriage return characters (i.e \textbackslash n and \textbackslash r), as they would interfere with the sentence alignment tool.

\subsection{Sentence alignment}

For sentence alignment, we used the LF aligner tool\footnote{\url{https://sourceforge.net/projects/aligner/}}, a wrapper around the Hunalign tool \cite{vargaparallel}, which provides an easy to use and complete solution for sentence alignment, including pre-loaded dictionaries for several languages. \par

Hunalign uses Gale-Church sentence-length information to first automatically build a dictionary based on this alignment. Once the dictionary is built, the algorithm realigns the input text in a second iteration, this time combining sentence-length information with the dictionary. When a dictionary is supplied to the algorithm, the first step is skipped. A drawback of Hunalign is that it is not designed to handle large corpora (above 10 thousand sentences), causing large memory consumption. In these cases, the algorithm cuts the large corpus in smaller manageable chunks, which may affect dictionary building.

The parallel abstracts were supplied to the aligner, which performed sentence segmentation followed by sentence alignment. A small modification in the sentence segmentation algorithm was performed to handle the fact that all words are in lowercase letters, which originally prevented segmentation. After sentence alignment, the following post-processing steps were performed: (i) removal of all non-aligned sentences; (ii) removal of all sentences with fewer than three characters, since they are likely to be noise.

\subsection{Machine translation evaluation}
To evaluate the usefulness of our corpus for SMT purposes, we used it to train an automatic translator with Moses \cite{moses_lang}. We also trained an NMT model using the OpenNMT system \cite{2017opennmt}, and used the Google Translate Toolkit\footnote{\url{https://translate.google.com/toolkit/}} to produce state-of-the-art comparison results. The produced translations were evaluated according to the BLEU score \cite{papineni2002bleu}.

\subsection{Manual evaluation}
\label{sub_manual_eval}
Although the Hunalign tool usually presents a good alignment between sentences, we also conducted a manual validation to evaluate the quality of the aligned sentences. We randomly selected 400 pairs of sentences. If the pair was fully aligned, we marked it as "correct"; if the pair was incompletely aligned, due to segmentation errors, for instance, we marked it as "partial"; otherwise, when the pair was incorrectly aligned, we marked it as "no alignment".

\section{Results and Discussion}
In this section, we present the corpus' statistics and quality evaluation regarding SMT and NMT systems, as well as the manual evaluation of sentence alignment.

\subsection{Corpus statistics}

Table \ref{table_stats} shows the statistics (i.e. number of documents and sentences) for the aligned corpus according to the 9 main knowledge areas defined by CAPES. The dataset is available\footnote{\url{https://figshare.com/articles/A_Parallel_Corpus_of_Thesis_and_Dissertations_Abstracts/5995519}} in TMX format \cite{Rawat2016}, since it is the standard format for translation memories. We also made available the aligned corpus in an SQLite database in order to facilitate future stratification according to knowledge area, for instance. In this database, we included the following metadata information: year, university, title in Portuguese, type of document (i.e. theses or dissertation), keywords in both languages, knowledge areas and subareas according to CAPES, and URL for the full-text PDF in Portuguese. An excerpt of the corpus is shown in Table \ref{table_excerpt} \par

\begin{table}[h]

\begin{center}
\begin{tabular}{|l|r|r|r|r|}
\hline  \multicolumn{1}{|c}{\bf Knowledge Area} & \multicolumn{1}{|c|}{\bf Docs} &\multicolumn{1}{c|}{\bf Sents} &\multicolumn{1}{c|}{\bf Tokens EN} &\multicolumn{1}{c|}{\bf Tokens PT}\\ \hline
Health Sciences          & 38,221  & 224,773 & 5.46M & 5.51M  \\
Humanities               & 38,493  & 189,648 & 5.63M & 5.54M  \\
Applied Social Sciences  & 32,176  & 160,131 & 4.66M & 4.60M  \\
Agricultural Sciences    & 26,740  & 154,710 & 3.92M & 3.92M  \\
Engineering              & 27,074  & 149,888 & 3.87M & 3.92M  \\
Multidisciplinary        & 26,502  & 140,849 & 3.84M & 3.81M  \\
Exact and Earth Sciences & 19,630  & 106,098 & 2.64M & 2.66M  \\
Biological Sciences      & 16,465  & 98,994  & 2.33M & 2.34M  \\
Linguistic and Arts      & 13,717  & 64,281  & 1.99M & 1.96M  \\
Total                    & 239,018 & 1,289,372 & 34.35M & 34.28M \\
\hline
\end{tabular}
\end{center}
\caption{\label{table_stats} Corpus statistics according to knowledge area. }

\end{table}

\begin{table}[!h]

\begin{center}
\begin{tabularx}{\linewidth}{|l|X|X|}
\hline  \multicolumn{1}{|c}{\bf ID} & \multicolumn{1}{|c|}{\bf Portuguese} &\multicolumn{1}{c|}{\bf English}\\ \hline
127454 & nessa tese apresentamos duas linhas de pesquisa distintas, a saber, na primeira, referente aos capítulos 1 e 3 aplicamos técnicas estatísticas à análise de imagens do satélite de abertura sintética (sar) e, na segunda, referente ao capítulo 2, examinamos problemas relativos à estimação de parâmetros por máxima verossimilhança na distribuição exponencial-poisson. & in this thesis we present two distinct research lines, namely, the first, referring to chapters 2 and 3, apply statistical techniques to the analysis of synthetic aperture radar (sar) images, and the second, referring to chapter 4, we examined problems concerning parameter estimation by maximum likelihood in exponential-poisson distribution. \\ \hline
    1419264 & para determinação dessa flora utilizamos os recursos de observação, coleta e identificação. & we use the resources of investigation, collection and identification to determine this flora. \\ \hline
    439358 & estimaram-se os benefícios ambientais da reciclagem de veículos com mais de 10 anos de uso, considerando os poluentes na fabricação de um veículo novo. & we estimated the environmental benefits of recycling vehicles in use more than 10 years, taking into consideration pollution engendered in the manufacture of a new vehicle. \\ \hline
    675023 & a coleta de dados se deu por meio de entrevista semiestruturada com 12 familiares cuidadores de crianças atendidas em pronto-socorro pediátrico de um hospital de ensino. & data collection was through semi-structured interviews with 12 family caregivers of children seen in a pediatric emergency department of a teaching hospital. \\ \hline
    675023 & os dados foram submetidos à análise de conteúdo temático conforme bardin (2011). & the data were subjected to thematic content analysis according to bardin (2011). \\ \hline
    1173306 & o planejamento e programação do projeto de construção naval têm dois objetivos por base: diminuir o tempo de fabricação e os custos. & shipbuilding project planning and scheduling possess two major objectives: manufacturing time and cost reduction. \\ 
\hline
\end{tabularx}
\end{center}
\caption{\label{table_excerpt} Excerpt of the corpus with document ID. }

\end{table}

\subsection{Translation experiments}
Prior to the MT experiments, sentences were randomly split in three disjoint datasets: training, development, and test. Approximately 13,000 sentences were allocated in the development and test sets, while the remaining was used for training. For the SMT experiment, we followed the instructions of Moses baseline system\footnote{\url{http://www.statmt.org/moses/?n=moses.baseline}}. For the NMT experiment, we used the Torch implementation\footnote{\url{http://opennmt.net/OpenNMT/}} to train a 2-layer LSTM model with 500 hidden units in both encoder and decoder, with 12 epochs. During translation, the option to replace UNK words by the word in the input language was used, since this is also the default in Moses. \par

Table \ref{table_BLEU} presents the BLEU scores for both translation directions with English and Portuguese on the development and test partitions for Moses and OpenNMT models. We also included the scores for Google Translate (GT) as a benchmark of a state-of-the-art system which is widely used. \par

\begin{table}[!h]
\begin{center}
\begin{tabular}{|c|c|c|c|}\hline
                \bf Partition    & \bf System           & \bf PT $\rightarrow$ EN     & \bf EN $\rightarrow$ PT    \\\hline
\multirow{2}{*}{Dev}  & Moses            & 44.07     &     41.21 \\
                      & OpenNMT          & 44.02     &     43.36 \\ \hline
\multirow{3}{*}{Test} & Moses            & 43.85     &     41.05 \\
                      & OpenNMT          & \bf 43.89 & \bf 43.22 \\
                      & Google Translate & 42.57     &     38.92 \\ \hline
\end{tabular}
\end{center}
\caption{\label{table_BLEU} BLEU scores for the translations using Moses, OpenNMT, and Google Translate. Bold numbers indicate the best results in the test set.}
\end{table}

NMT model achieved better performance than the SMT one for EN $\rightarrow$ PT direction, with approximately 2.17 percentage points (pp) higher, while presenting almost the same score for PT $\rightarrow$ EN. When comparing our models to GT, both of them presented better BLEU scores, specially for the 
EN $\rightarrow$ PT direction, with values ranging from 1.27 pp to 4.30 pp higher than GT. \par

We highlight that these results may be due to two main factors:  corpus size, and domain. Our corpus is fairly large for both SMT and NMT approaches, comprised of almost 1.3M sentences, which enables the development of robust models. Regarding domain, GT is a generic tool not trained for a specific domain, thus it may produce lower results than a domain specific model such as ours. Scientific writing usually has a strict writing style, with less variation than novels or speeches, for instance, favoring the development of tailored MT systems. \par

Below, we demonstrate some sentences translated by Moses and OpenNMT compared to the suggested human translation. One can notice that in fact NMT model tend to produce more fluent results, specially regarding verbal regency.

\smallskip

\begin{quote}
Human translation: \textit{this paper presents a study of efficiency and power management in a packaging industry and plastic films.}
\end{quote}

\begin{quote}
OpenNMT: \textit{this work presents a study of efficiency and electricity management in a packaging industry and plastic films.}
\end{quote}

\begin{quote}
Moses: \textit{in this work presents a study of efficiency and power management in a packaging industry and plastic films.}
\end{quote}

\begin{quote}
GT: \textit{this paper presents a study of the efficiency and management of electric power in a packaging and plastic film industry.}
\end{quote}
\smallskip

\begin{quote}
Human translation: \textit{this fact corroborates the difficulty in modeling human behavior.}
\end{quote}

\begin{quote}
OpenNMT: \textit{this fact corroborates the difficulty in modeling human behavior.}
\end{quote}

\begin{quote}
Moses: \textit{this fact corroborated the difficulty in model the human behavior.}
\end{quote}

\begin{quote}
GT: \textit{this fact corroborates the difficulty in modeling human behavior.}
\end{quote}

\smallskip

\subsection{Sentence alignment quality}

We manually validated the alignment quality for 400 sentences randomly selected from the parsed corpus and assigned quality labels according Section \ref{sub_manual_eval}. From all the evaluated sentences, 82.30\% were correctly aligned, while 13.33\% were partially aligned, and 4.35\% presented no alignment. The small percentage of no alignment is probably due to the use of Hunalign tool with the provided EN/PT dictionary. \par

Regarding the partial alignment, most of the problems are result of segmentation issues previous to the alignment, which wrongly split the sentences. Since all words were case folded to lowercase letters, the segmenter lost an important source of information for the correct segmentation, generating malformed sentences. Some examples of partial alignment errors are shown in Table \ref{table_errors}, where most senteces were truncated in the wrong part. 
\begin{table}[!h]

\begin{center}
\begin{tabularx}{\linewidth}{|X|X|}
\hline  \multicolumn{1}{|c|}{\bf Portuguese} &\multicolumn{1}{c|}{\bf English}\\ \hline
os dados foram comparados entre os grupos por anova de medidas repetida & data were compared by repeated measures anova. results: we found a significa \\ \hline
o estudo utilizará um software comercial para simular a peça & the study will use commercial software to simulate the piece with a number of different crack sizes and the \\ \hline

buscamos subsídios teóricos em autores que veem na reflexão e na pesquisa um grande potencial para o desenvolvimento d	& we seek theoretical support in authors who see in reflection and research a great potential for \\

\hline
\end{tabularx}
\end{center}
\caption{\label{table_errors} Examples of partial alignment errors. }

\end{table}

\section{Conclusion and future work}

We developed a parallel corpus of theses and dissertations abstracts in Portuguese and English. Our corpus is based on the CAPES TDC dataset, which contains information regarding all theses and dissertations presented in Brazil from 2013 to 2016, including abstracts and other metadata. \par

Our corpus was evaluated through SMT and NMT experiments with Moses and OpenNMT systems, presenting superior performance regarding BLEU score than Google Translate. The NMT model also presented superior results than the SMT one for the EN $\rightarrow$ PT translation direction. We also manually evaluated sentences regarding alignment quality, with average 82.30\% of sentences correctly aligned. \par

For future work, we foresee the use of the presented corpus in mono and cross-language text mining tasks, such as text classification and clustering. As we included several metadata, these tasks can be facilitated. Other machine translation approaches can also be tested, including the concatenation of this corpus with other multi-domain ones.

%
%
%
\bibliographystyle{splncs04}
\bibliography{main}

\end{document}